\documentclass[sigconf]{acmart}

\AtBeginDocument{%
  \providecommand\BibTeX{{%
    \normalfont B\kern-0.5em{\scshape i\kern-0.25em b}\kern-0.8em\TeX}}}

\setcopyright{none}
\renewcommand\footnotetextcopyrightpermission[1]{}
\begin{document}

\title{Controllable Longer Image Animation with Diffusion Models}


\author{Qiang Wang}
\affiliation{%
  \institution{Alibaba Group}
  \city{Beijing}
  \country{China}}
\email{yijing.wq@alibaba-inc.com}

\author{Minghua Luo}
\affiliation{%
  \institution{Alibaba Group}
  \city{Beijing}
  \country{China}}
\email{luominghua.lmh@alibaba-inc.com}

\author{Junjun Hu}
\affiliation{%
  \institution{Alibaba Group}
  \city{Beijing}
  \country{China}}
\email{hujunjun.hjj@alibaba-inc.com}

\author{Fan Jiang}
\affiliation{%
  \institution{Alibaba Group}
  \city{Beijing}
  \country{China}}
\email{frank.jf@alibaba-inc.com}

\author{Mu Xu}
\affiliation{%
  \institution{Alibaba Group}
  \city{Beijing}
  \country{China}}
\email{xumu.xm@alibaba-inc.com}

\begin{abstract}
Generating realistic animated videos from static images is an important area of research in computer vision. Methods based on physical simulation and motion prediction have achieved notable advances, but they are often limited to specific object textures and motion trajectories, failing to exhibit highly complex environments and physical dynamics. In this paper, we introduce an open-domain controllable image animation method using motion priors with video diffusion models. Our method achieves precise control over the direction and speed of motion in the movable region by extracting the motion field information from videos and learning moving trajectories and strengths. Current pretrained video generation models are typically limited to producing very short videos, typically less than 30 frames. In contrast, we propose an efficient long-duration video generation method based on noise reschedule specifically tailored for image animation tasks, facilitating the creation of videos over 100 frames in length while maintaining consistency in content scenery and motion coordination. Specifically, we decompose the denoise process into two distinct phases: the shaping of scene contours and the refining of motion details. Then we reschedule the noise to control the generated frame sequences maintaining long-distance noise correlation. We conducted extensive experiments with 10 baselines, encompassing both commercial tools and academic methodologies, which demonstrate the superiority of our method. Our project page: \url{https://wangqiang9.github.io/Controllable.github.io/}
\end{abstract}




\maketitle
\begin{figure*}[h]
  \centering
  \includegraphics[width=\linewidth]{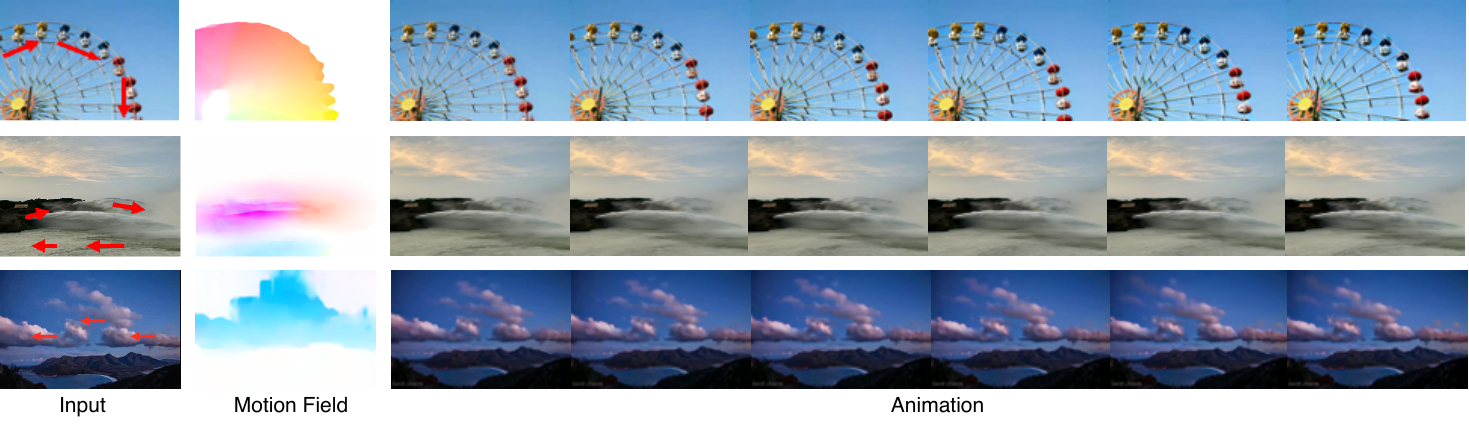}
  \caption{Examples of our method for image animation. The first column displays the input reference image in conjunction with the arrow controls, serving as motion control. The second column depicts the refined motion field based on the directional information provided by the input arrows. The final column showcases selected frames from the generated animation sequence, specifically frames 4, 8, 12, 16, 20 and 24.}
  \label{fig:fig1}
\end{figure*}
\section{Introduction}
Image animation has always been a task of great interest in the field of computer vision, with the goal of converting still images into videos that conform to the laws of motion. Videos featuring natural and vivid motion are significantly more appealing than static images, leading to their extensive use in film production, advertising, social networking, and various other sectors. Nevertheless, creating such videos presents considerable challenges. Previous works predominantly concentrated on the neural aspects \cite{2019Animating} or physical attributes \cite{chuang2005animating} of the surface texture, which mainly focusing on scenes within specific domains, such as natural scenes \cite{fan2023simulating, mahapatra2022controllable, holynski2021animating}, portraits \cite{geng2018warp, wang2020imaginator, wang2022latent}, bodies \cite{bertiche2023blowing, blattmann2021understanding, karras2023dreampose}, which limits their application to complex motions in a wide range of real-world scenarios. 

Recently, diffusion models \cite{rombach2022high, nichol2021glide, ramesh2022hierarchical} trained on expansive datasets have achieved remarkable progress in the generation of images that are both high-quality and diverse. Encouraged by this success, researchers extended these models to the realm of video generation \cite{chen2024videocrafter2, blattmann2023align, ho2022video, ho2022imagen} by leveraging the strong image generative priors, making it possible to generate realistic and diverse videos. However, due to the reliance on text and reference images as conditions, these models lack precise control over object motion when faced with the challenge of complex spatio-temporal prior modeling. On the other hand, most existing base models can only reason about models below 30 frames due to the scarcity of high-quality long video datasets and limitations of computation resources \cite{wang2023gen}.

In this paper, we aim to design a controllable longer image animation model that can address such problems. We separately consider the motion in the video into the object's motion and the overall scene movement. Specifically, for the object's motion, we extract the motion field of its trajectory and impose constraints on both direction and speed. This enables users to precisely control the detailed trajectory of the object using sparse trajectory methods, such as arrows. For the overall scene movement, we calculate the intensity embedding of the entire motion field to control the motion strength. Our method overcomes limitations of traditional approaches that are domain-specific, enabling precise control of motion in an open-domain setting. Additionally, we examine the principles of motion reconstruction during the denoise process and introduce a phased inferencing strategy predicated on shared noise variables. Through the decomposition of scene contour and motion detail inference, the consistency of temporal features are maintained and the artifacts and flickers are reduced. Furthermore, the cost of longer animation inference is significantly reduced thanks to the phased inference method. We summarize the contributions of this paper as follows:
\begin{itemize}
    \item We propose a controllable method for generating videos from static images. Our approach imposes fine-grained constraints on the motion of moving targets in videos by utilizing optical flow fields, controlling the direction, speed and strength of movement.
    \item We investigate the relationship between video consistency preservation and noise during the denoise process, and propose a method for generating long videos based on shared noise reschedule that yields better visual effects.
    \item Our method overcomes the shortcomings of previous methods that were limited to particular domains, and achieves the highest quality of generative results across multiple benchmark when compared with various methods.
\end{itemize}
\setlength{\itemsep}{1ex} 

\fancyhead{} 
\fancyfoot{} 

\section{Related Work}
\subsection{Image Animation}
\thispagestyle{empty}
Image animation is a challenging task, and early works relied on physical simulations \cite{chuang2005animating, jhou2015animating} and motion predictions \cite{2019Animating, mahapatra2022controllable, geng2018warp, wang2020imaginator, wang2022latent}. Methods utilizing physical simulation emulate object movements using physics principles, exemplified by the oscillation of a sailboat upon the sea.  Such approaches necessitate precise knowledge of each object's identity and motion scope, as well as straightforward and replicable motion principles, rendering them inapplicable to general scenarios. Methods based on motion prediction employ recursive motion prediction or motion field prediction to model object movements. Methods utilizing motion prediction \cite{2019Animating} lead to the gradual accumulation of errors and results in distortions when creating successive video clips. To overcome this issue, motion  
fields prediction methods \cite{fan2023simulating, mahapatra2022controllable, hao2018controllable, holynski2021animating} adopt motion estimation networks to guide the movement of objects. Holynski et al. \cite{holynski2021animating} leverage a single motion estimation and static Eulerian flow fields to depict the motion information of images at different moments, using warping of the flow field with image features to generate subsequent frames. 
Mahapatra et al. \cite{mahapatra2022controllable} achieve control over the direction of fluid motion and the motion of specific elements by converting arrows and masks into optical flow representations. Similarly, Hao et al. \cite{hao2018controllable} adopt the concept of transforming sparse motion trajectories into dense optical flow for meticulous motion manipulation. However, the majority of these approaches are primarily concentrated on the motion of textured surfaces of objects, such as flowing water, and fail to extend to the motion of rigid bodies, like Ferris wheels or tree branches. This constraint hampers their versatility and scalability.

\subsection{Diffusion Models}
Diffusion models (DMs) \cite{ho2020denoising, song2020denoising} have recently shown better sample quality, stability and conditional generation capabilities than Variational Autoencoders(VAEs) \cite{kingma2013auto}, GANs \cite{goodfellow2020generative}, and Flow Models \cite{dinh2014nice}. Prafulla Dhariwal et al. \cite{dhariwal2021diffusion} demonstrated that DMs can achieve state-of-the-art(SOTA) results in terms of image sampling quality guided by classifiers. Stable Diffusion \cite{rombach2022high} has shown unprecedented capabilities in image generation through denoise diffusion in the latent space, using language condition\cite{radford2021learning}. To incorporate additional conditions for constraining the generated results, ControlNet \cite{zhang2023adding} and T2I-Adapter \cite{mou2023t2i} add new modules to accept additional image inputs, achieving precise generative layout control. Composer \cite{huang2023composer} controls generation using shape, global information and color histogram as the local guidance. Many fine-tuning strategies such as LoRA \cite{hu2021lora} , DreamBooth \cite{ruiz2023dreambooth} are also developed to force DMs with different new concepts and styles. Building upon these controllable developments in DMs, our work leverages motion information based on optical flow fields as a controlling condition, providing fine-grained guidance for diffusion models.


\subsection{Video Generation}

GANs \cite{goodfellow2020generative} and Transformers \cite{vaswani2017attention} are the commonly used backbones in early research of video generation, e.g., StyleGAN-V\cite{skorokhodov2022stylegan}, VGAN \cite{vondrick2016generating}, TGAN\cite{saito2017temporal}, MoCoGAN\cite{tulyakov2018mocogan}, VideoGPT\cite{yan2021videogpt},  MAGVIT \cite{yu2023magvit} and NUVA-infinity \cite{wu2022nuwa}. Recently, inspired by the significant advancements in image generation, DMs-based video generation methods have made significant progress\cite{ho2022imagen, luo2023videofusion, chen2024videocrafter2}. Most methods of the backbone is a 3D UNet \cite{blattmann2023align} with time aware capabilities that directly generates complete video blocks. AnimatedDiff\cite{guo2023animatediff} extends Stable Diffusion \cite{rombach2022high} by only training added temporal layers within a 2D UNet, which can be combined with the weights of a personalized text-to-image model. VideoComposer \cite{wang2024videocomposer} introduces a spatio-temporal condition encoder that flexibly synthesizes videos while enhancing frame sequence consistency. I2VGen-XL\cite{2023i2vgenxl} decouples the generation process by adopting a decomposition approach, preserving high-level semantics and low-level details through two encoders. LFDM \cite{ni2023conditional} leverages the spatial position of a given image and learns a low-dimensional latent flow space based on temporally-coherent flow to control synthetic video. While time-aware dilated UNet schemes maintain a fixed temporal resolution throughout the network, most clip frames are limited to within 30 frames, long lengths clips are restricted. Our contribution lies in the context of still image animation, controlling the details of motion through optical flow and enabling longer video sequence prediction.
\begin{figure*}[h]
  \includegraphics[width=\linewidth]{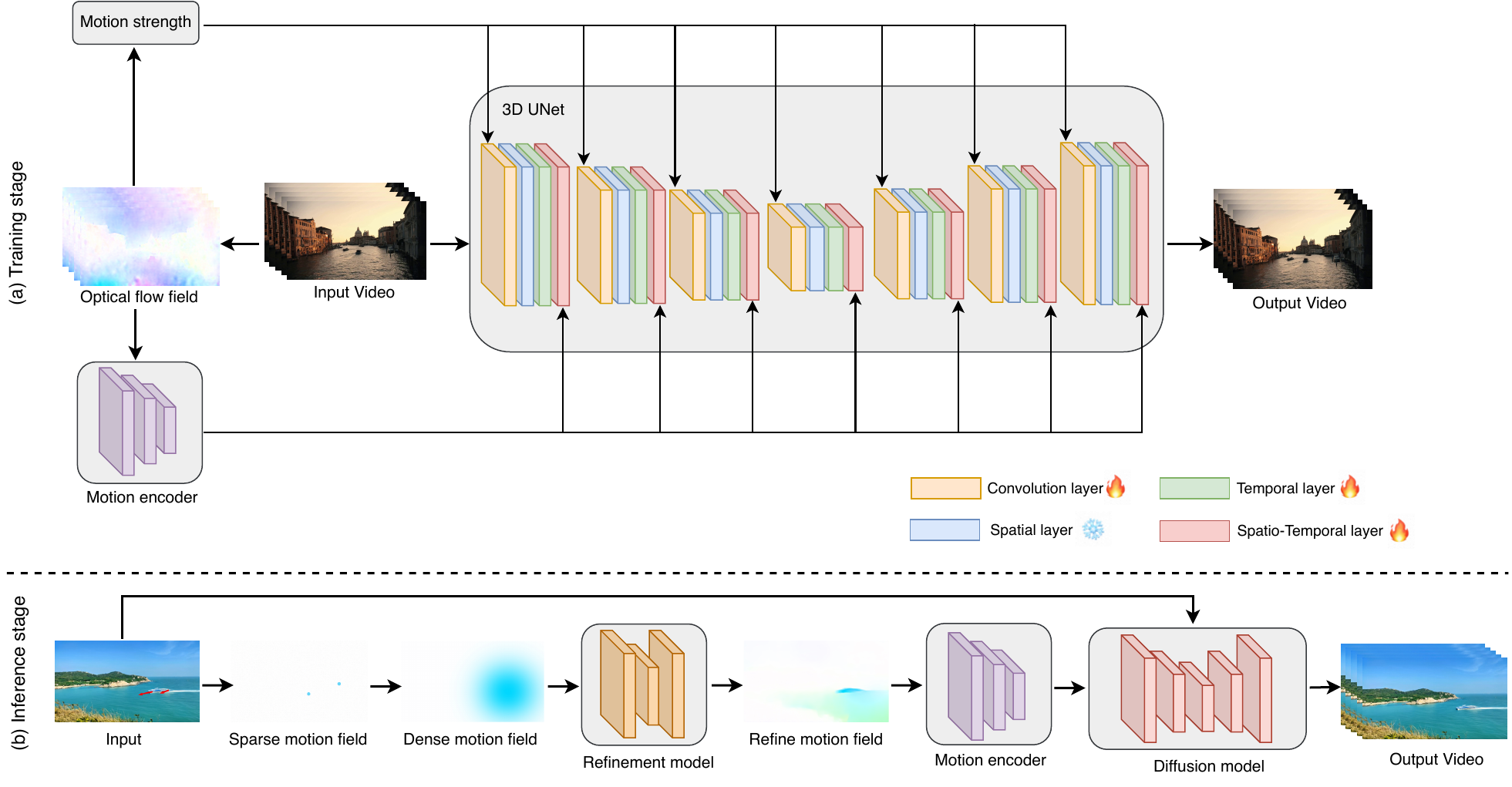}
  \caption{Overview of motion fields guidance: (a) Training stage: We extract optical flow motion field and motion strength from training videos as conditional constraints. The motion field is enhanced through a spatio-temporal layer attention mechanism, while the motion intensity is projected into positional embeddings and concatenated with timestep embeddings. (b) Inference stage: The control arrow provided by the user is initially transformed into a sparse motion field, and then convert to dense motion field by interpolation. Subsequently, the refined motion field is produced by employing a refinement model. The motion field, in conjunction with the input motion strength, regulates the video generation.}
  \label{fig:overview}
\end{figure*}
\section{Method}
Given a reference image $I_0$, our target is to generate a sequence of subsequent video frames {$ \left\{\hat{I}_1, \hat{I}_2, \cdot, \hat{I}_N \right\}$}. As shown in Fig. \ref{fig:overview} , we extract motion field information (detailed in Sec. \ref{subsec:motion_fields}) and exercise intensity information (detailed in Sec. \ref{subsec:moiton_strength}) to guide the generation process. We analyzed the characteristics of different noise levels and denoise stages, and proposed a longer video generation method based on phased inference and shared noise reschedule in Sec. \ref{subsec:longer}. We start with introducing the preliminary knowledge of the latent diffusion models(LDMs)\cite{rombach2022high} in Sec. \ref{subsec:Preliminaries}.

\thispagestyle{empty}

\subsection{Preliminaries} \label{subsec:Preliminaries}

We choose LDMs\cite{rombach2022high} as the generative model backbone, which utilize a pre-trained VAEs \cite{kingma2013auto} to encode video data $x_0 \in \mathbb{R}^{L \times C \times H \times W}$ frame-by-frame, where respectively $L$, $C$, $H$, and $W$ denote the video length, number of channels, height, and width. After encoding, we obtain latent representation $z_0 \in \mathbb{R}^{L \times c \times h \times w}$. The forward process of LDMs is a Markov process that slowly iterates to inject Gaussian noise $\epsilon$, disrupting the data distribution to obtain $z_t$ for each timestep $t$, where  $t = 1, \ldots, T$ and $T$ denotes the total number of timesteps:
\begin{equation}
\label{eq:zt}
  z_t=\sqrt{\bar{\alpha}_t} z_0+\sqrt{1-\bar{\alpha}_t} \epsilon, \quad \epsilon \sim \mathcal{N}(0, I),
\end{equation}

where $\bar{\alpha}_t=\prod_{i=1}^t\left(1-\beta_t\right)$ with $\beta_t$ is the noise intensity coefficient in timestep $t$. During training, the noise prediction function $\epsilon_\theta$ is trained to predict $\epsilon$ using the following mean-squared error loss:
\begin{equation}
\label{eq:epsilon}
    l_\epsilon=\left\|\epsilon-\epsilon_\theta\left(z_t, t, c\right)\right\|_2^2,
\end{equation}

where $c$ is the user-input condition to control the generation process flexibly. In this paper, we choose Stable Video Diffusion \cite{blattmann2023stable} as the base LDMs. The noise predictor $\epsilon_\theta$ is implemented as a 3D UNet \cite{blattmann2023align} architecture, which is constructed with a sequence of blocks that consist of convolution layers, spatial layers, temporal layers and spatio-temporal layers (show in Fig. \ref{fig:overview}). Considering that our discussion on the motion portrayal of static images, the camera remains stationary. We train the static camera motion LoRA\cite{hu2021lora} within the temporal layers \cite{blattmann2023align, guo2023animatediff}.

\subsection{Motion Fields Guidance} \label{subsec:motion_fields}

We introduce motion fields guidance to provide users with precise control over the movable areas of the input image. As shown in Fig. \ref{fig:overview}, we first convert video clips into optical flow sequences. Subsequently, we employ the motion encoder to extract motion guidance condition $c$, which is incorporated into spatial-temporary cross attention in $\epsilon_\theta$. 

\noindent {\bf{Motion Fields Estimation.}}~ The motion of adjacent pixels in the video is similar, which is suitable for expressing the motion between video frames by the optical flow field. Given two consecutive images from videos, we adopt RAFT \cite{teed2020raft} to estimate the optical flow field. The optical flow of the $k$-th frame on each pixel coordinate $(x_k, y_k)$ can be expressed as a dense pixel displacement field $F_{k \rightarrow k+1}=(f^x_{k \rightarrow k+1}, f^y_{k \rightarrow k+1})$. The coordinates of each pixel in the next (${k+1}$)-th frame can be represented by the displacement field projection as:
\begin{equation}
\label{eq:fk}
    (x_{k+1}, y_{k+1}) = (x_k, y_k) + F_{k \rightarrow k+1},
\end{equation}

Motion Fields $\mathbb F $ refers to a collection of optical flows consisting of N frames $\{F_{0 \rightarrow 1}, F_{1 \rightarrow 2}, \ldots, F_{N-1 \rightarrow N}\}$. Follow the work of Endo Y.et al. \cite{2019Animating}, we adopt the CNN-based motion encoder to transform $\mathbb F$ into the motion feature maps $z_m$. Subsequently, we calculate the cross-attention value of the latent feature $z$ in the spatio-temporal attention layer:
\begin{equation}
\label{eq:attention}
    Attention(Q, K, V_m) = Softmax(\dfrac {Q K^T} {\sqrt{d}}) V_m
\end{equation}

where $\sqrt{d}$ is a scaling factor, and $Q=W^Q z$, $K=W^K z$, $V_m = W^V z_m$ are projection operation. Due to the spatio-temporal attention layer needing to consider both appearance and motion information simultaneously, our method increases the model's receptive fields for the motion information.

\noindent {\bf{Sparse Trajectory Control.}}~ Sparse trajectories (such as arrows) are easier to interact and express. Specifically, users input arrows $A^{1:S}$ and object motion strengths $M_o^{1:S}$ to control the direction and speed of object motion, accurately specifying the expected motion of the target pixel. The starting point $(x_i, y_i)$ to the ending point $(x_j, y_j)$ of each arrow $A^s$ in the input image. Inspired by the work of Hao et al. \cite{hao2018controllable}, we first convert them into sparse optical flow motion fields $f_s$:
\begin{equation}
\label{eq:FS}
    f_s\left(x^i, y^i\right)= \begin{cases}\left(x^j, y^j\right) * M_o^s & \text { if } A^s \text { starts at }\left(x^i, y^i\right) \\ 0 & \text { otherwise }\end{cases}
\end{equation}

As observed in Fig. \ref{fig:overview}, there is a significant difference in the density of the optical flow between the sparse optical flow field and the actual optical flow. Therefore, we use $f_s$ to generate a dense optical flow motion field, $f_d$. Inspired by the work of A Mahapatra et al. \cite{mahapatra2022controllable}, we perform weighted interpolation on the motion field near the sparse optical flow field, with the range of interpolation limited by a threshold $R$:
\begin{equation}
\label{eq:fd_hat}
     \hat f_d\left(x^j, y^j\right) = \sum_{i=1}^N  e^{{-(D/\sigma)}^2} * f_s(x^i, y^i)
\end{equation}

\begin{equation}
\label{eq:Fd}
     f_d\left(x^j, y^j\right)= \begin{cases}\hat f_d\left(x^j, y^j\right) & \text { if } \hat f_d >  R  \\ 0 & \text { otherwise }\end{cases}
\end{equation}

where $D$ is the Euclidean distance between $(x^i, y^i)$ and $(x^j, y^j)$, and $\sigma$ is a hyper parameter proportional to frame size. However, this motion description field can only provide a rough temporal trend, which differs from the established optical flow field in the train stage. Therefore, additional methods are required to refine and correct this motion description. Drawing inspiration from the research conducted by P Isola et al. \cite{isola2017image}, our study develops and constructs a pixel-to-pixel refinement model $T$, which transitions from a dense optical flow field to a refined optical flow field. This advancement corrects the depiction of object motion and improves the model's capacity to discern and capture the subtleties of movements.

\subsection{Motion Strength Guidance} \label{subsec:moiton_strength}

We introduce object motion strength guidance in Sec \ref{subsec:motion_fields}. However, controlling local object motion is inadequate. We propose global motion strength condition $M_s$ to manage the intensity of motion in the entire scene, especially for the scene background, and calculate the arithmetic mean of the absolute value of the motion fields: 
\begin{equation}
\label{eq:strenth_s}
    M_s = \dfrac {\sum^{N-1}_{k=0} |F_{k \rightarrow k+1}|} {N}
\end{equation}

The global motion strength quantitatively measures how great the motion intensity is between each frame. We project $M_s$ into a positional embedding and then concatenate it with timesteps embeddings before feeding it into the convolution layers.

\subsection{Longer Video Generation } \label{subsec:longer}

To generate longer animation using video diffusion models, a straightforward solution is to directly take the last frame of the previous video as condition input for the next video and iteratively infer and concatenate multiple short videos $V^k$ into the longer video $\{V^1, \ldots, V^K\}$, where $K$ is the total number of short videos. Unfortunately, as shown in Fig. \ref{fig:longer_compare}, it generates inconsistent motion and temporal jittering between the spliced videos. Considering the characteristic of images animation, we propose a method of shared noise consistency phased inference based on peculiarity in different inference stages, which synthesizes longer videos with better consistency and significantly reduces inference costs.
\begin{figure}[h]
  \centering
  \includegraphics[width=\linewidth]{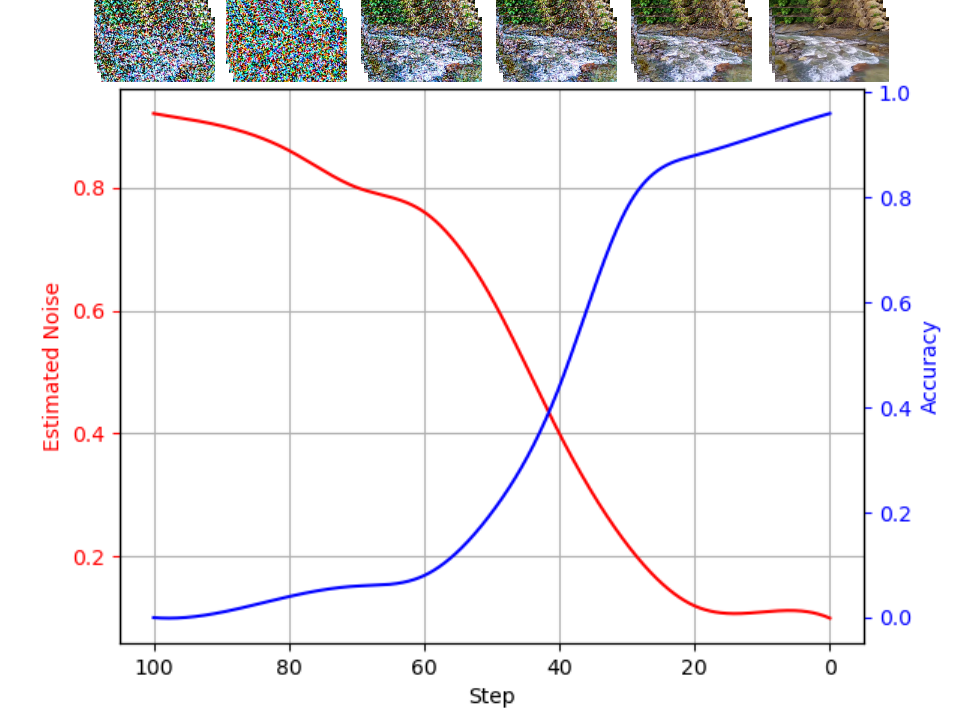}  \caption{Variability in noise patterns and contour accuracy is evident across different timesteps. The upper part of the curve graph illustrates the visual outcomes at every 20-step interval.}
  \label{fig:longer_theory}
\end{figure}

\thispagestyle{empty}

\begin{figure*}[h]
  \includegraphics[width=\linewidth]{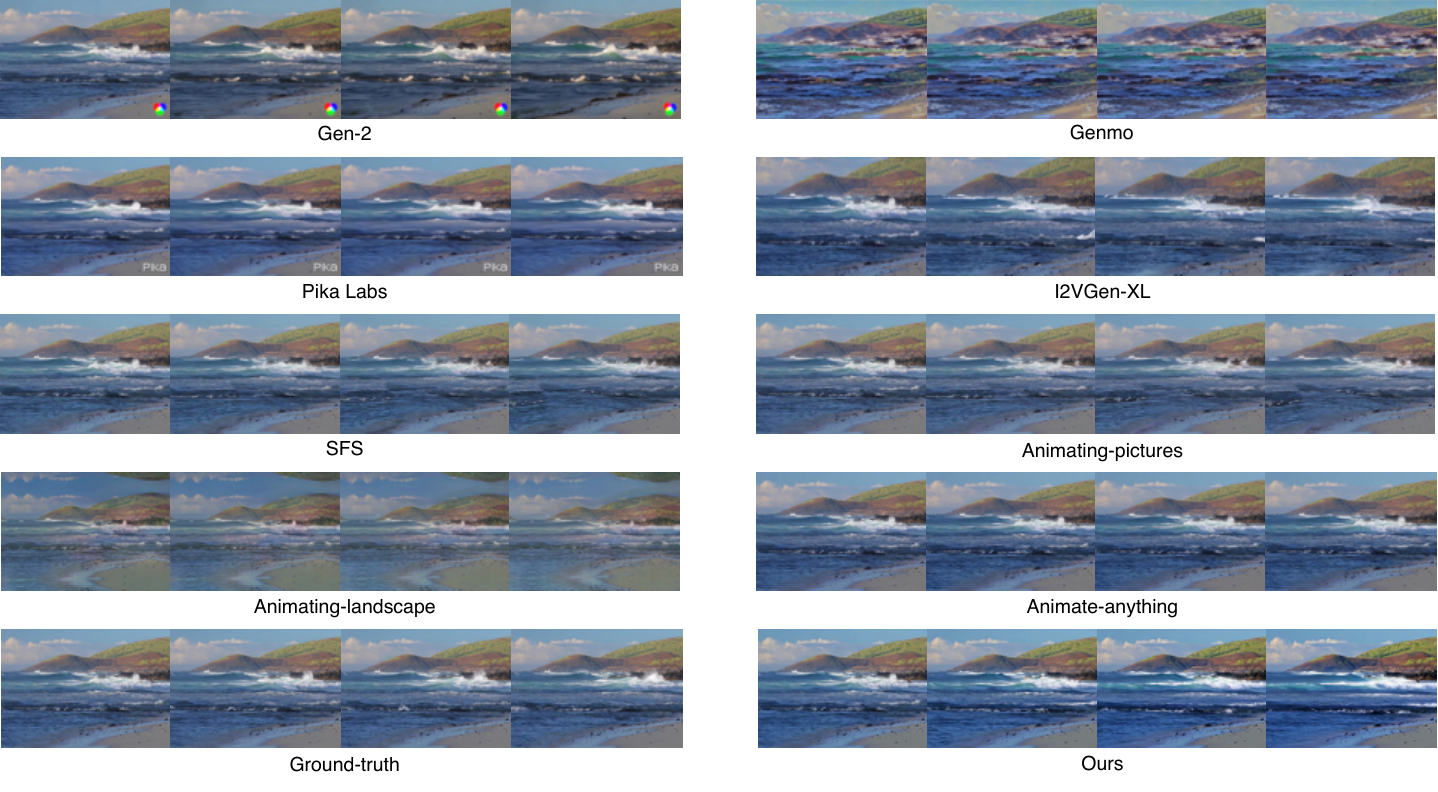}
  \caption{Qualitative results between baselines and our approach. Additional examples are provided in the supplemental material.}
  \label{fig:compare_all}
\end{figure*}
\noindent {\bf{Phased Inference.}}~ During the denoise process, the contributions of different stages to the final outcome are imbalanced \cite{wang2022sketchknitter}. We found a similar regularity in the process of video sampling. As shown in Fig. \ref{fig:longer_theory}, we determine accuracy by the degree of similarity in contour segmentation \cite{kirillov2023segment} between frames at different timesteps and the resultant frame. Additionally, we gauge the estimated noise patterns by considering the mean value of the noise, estimated noise patterns exhibit variation across distinct timesteps. Considering that the camera shots in image animation are fixed, the main appearances and contours in the videos are determined in the early stage of inference, while the detailed motion is formed in the late stage. Therefore, we carry out staged inference with multiple short videos, the denoise process of $V^1$ is complete, and $V^{2:K}$ only needs to resample the formation of detail motion based on $V^1$.
\begin{equation}
\label{eq:phased}
    \hat V^k =
    \begin{cases} \mathbb D \left(z^k_{1:T}\right) & k=1 \\
    \mathbb D \left(z^1_{1:M}, z^k_{(M+1):T}\right) & k>1 
    \end{cases}
\end{equation}

where $M = {\lfloor \gamma \cdot T \rfloor}$, ${\lfloor  \rfloor}$ is upward rounding symbol,  $\gamma$ is defined as the hyperparameter that adjusts the segmentation inference phase, and $\mathbb D$ is denoise operation. Since this approach eliminates numerous redundant inference steps, it can significantly reduce the inference time.

\noindent {\bf{Shared Noise Reschedule.}}~ We added a certain number of steps of noise to the input image, obtaining a noise prior containing information of the input image. According to Eq. \ref{eq:zt}, we use $\epsilon_\theta$ to predict the noise $n(t)$ at the $t$ timestep:
\begin{equation}
\label{eq:nt}
    n(t) = \epsilon_\theta(z_t, t, c), 
\end{equation}

This approach achieves a balance between retaining image features and noise consistency. However, we observed that this limits the motion diversity of the generated video segments, thus we introduced more randomness into the new noise $\tilde n(t)$:
\begin{equation}
\label{eq:nt_tilde}
    \tilde n(t) = n(t) + \omega \cdot \epsilon, \quad \epsilon \sim \mathcal{N}(0, I), 
\end{equation}
where $\omega$ is is the hyper parameter for adjusting the level of randomness, and $t \geq M$. In order to maintain consistent noise correlation between multiple video segments, $\tilde n(t)^{0:L-1}_k$ is the $k$-th noise sequence for each video segment with length $L$, and the noise sequence for the long video is:
\begin{equation}
\label{eq:long_noise}
    [\tilde n(t)^{0:L-1}_2, \tilde n(t)^{0:L-1}_3, \ldots, \tilde n(t)^{0:L-1}_K],
\end{equation}

\thispagestyle{empty}

Then we randomly shuffle it, maintaining the remote correlation and randomness of noise in all short video clips.

\section{Experiments}

\subsection{Experimental Setup}

\noindent {\bf{Dataset.}}~ We randomly selected $30,000$ videos from HDVILA-100M \cite{xue2022advancing} as the training dataset, and filtered $5,000$ videos with fixed lenses to train the static camera motion module. The initial frame of each video is utilized as the input image condition. In order to train the optical flow field refinement model $T$, we utilized the training dataset introduced by A Holynski et al. \cite{holynski2021animating}. The dataset comprises approximately $5,000$ training videos, each accompanied by a detailed annotation of the refined optical flow field.

To ensure a comprehensive and impartial assessment, we have established a benchmark tailored for quantitative analysis. We downloaded $1024$ videos from the copyright-free website Pixabay, including various categories such as natural scenery, amusement parks, etc. Each video was truncated from its start into two distinct lengths: brief clips consisting of $16$ frames, and extended sequences of $125$ frames. This longer sequences is evaluated our longer image animation algorithm. The results of all methods were truncated to the same frame length.

\noindent {\bf{Evaluation Metrics.}}~ To evaluate our paradigm, we evaluated the quality of results by Fréchet Video Distance({\bf{FVD}}) \cite{unterthiner2018towards}, Peak Signal-to-Noise Ratio({\bf{PSNR}}) , Structural Similarity ({\bf{SSIM}}) \cite{wang2004image}, Learned Perceptual Image Patch Similarity ({\bf{LPIPS}}) \cite{zhang2018unreasonable} and Temporal Consistency ({\bf{Tem-Cons}}). FVD is used to measure visual quality, temporal coherence, and diversity of samples, PSNR and SSIM is used to measure the frame quality at the pixel level and the structural similarity between synthesized and real video frames. LPIPS is used to measure the perceptual similarity between synthesized and real video frames. Tem-Cons evaluates the temporal consistency of a video by calculating the average cosine similarity within the CLIP \cite{radford2021learning} embedding space across adjacent frames. Furthermore, we conducted a use study from two different aspects: Motion Coordination({\bf{Mo-Coor}}), Visual Coherence({\bf{Vi-Co}}), Overall Aesthetics({\bf{Overall-Aes}}). Respectively measure the adherence of generated video motion to physical laws and the preservation of picture consistency, the coherence and coordination of object movement and aesthetic appeal with respect to conditional image input as perceived by humans. The experiment involved the selection of $100$ generated samples, which were then combined with the other three baseline generated videos. A total of $32$ individuals took part in this experiment, each participant was tasked with selecting the video that displayed the most consistent with the individual's rating criteria.


\noindent {\bf{Implements Details.}}~ We employ the AdamW optimizer \cite{loshchilov2017decoupled} with a constant learning rate of $3 \times 10^{-5}$ for training the video diffusion model and $5 \times 10^{-3}$ for training the optical flow field refinement model. We freeze LDM autoencoder to individually encode each video frame into latent representation. The spatial layer of the UNet is set to a frozen state, while the remainder of the model remains trainable. The model is trained using a dataset of $25$ frames and is expected to generate samples from $125$ frames during the inference stage. The training videos have a resolution of $512 \times 512$. The training of static camera motion LoRA is conducted separately, with the LoRA rank set to 16. The phased inference hyperparameter $\gamma$ is $0.8$ and randomness level $\omega$ is $0.2$. During the training of motion LoRA, other parts of the model are frozen. For controlling the sparse trajectory, the interpolation threshold $R$ is configured to $0.05$, and $\sigma$ is established at $170$. All experiments are conducted on on two NVIDIA A100 GPUs. Our training regimen for the diffusion models consisted of $100,000$ iterations with a batch size of $4$.  This training process required around $20$ hours to complete. We trained the optical flow refinement models over $10,000$ iterations, employing a batch size of $32$, which culminated in roughly $8$ hours.

\thispagestyle{empty}

\subsection{Comparison with baselines} 

\begin{table*}
    \caption{Quantitative results. The metrics corresponding to the top-performing method are accentuated in \textcolor{red}{red}, whereas those representing the second most effective approach are underscored in \textcolor{blue}{blue}.}
    \label{tab:qualitative_results}
    \begin{tabular}{l|ccccc|ccc}
    \hline & \multicolumn{5}{|c}{ Automatic Metrics} & \multicolumn{3}{c}{User Study} \\
    Method & FVD $\downarrow$ & PSNR $\uparrow$ & SSIM $\uparrow$ & LPISP $\downarrow$ & Tem-Cons$\uparrow$ & Mo-Coor $\uparrow$ & Vi-Co $\uparrow$ & Overall-Aes $\uparrow$ \\
    \hline \hline Gen-2 \cite{gen-2} & 312.66 & 18.39 & 0.5542 & 0.2941 & 0.9375 & 88.76 & 86.75 & {90.51} \\
    Genmo \cite{Genmo} & 432.24 & 14.13 & 0.1854 & 0.3090 & \textcolor{blue}{0.9772} & 82.98 & 85.87 & 78.14 \\
    Pika Labs \cite{Pika} & {304.78} & 18.81 & 0.6144 & 0.3130 & 0.9566 & 84.17 & \textcolor{blue}{90.22} & 86.79 \\
    I2VGen \cite{2023i2vgenxl} & 301.81 & 19.48 & 0.6205 & 0.2663 & 0.9598 & 91.54 & 88.13 & 86.65 \\
    SFS \cite{fan2023simulating} & {288.03} & \textcolor{red}{22.51} & 0.6176 & {0.1697} & 0.9715 & \textcolor{blue}{93.73} & 88.96 & 85.23 \\
    Animating-pictures \cite{holynski2021animating} & 294.65 & 21.50 & 0.6214 & 0.1632 & 0.9682 & 90.74 & 83.45 & 86.89 \\
    Animating-landscape \cite{2019Animating} & 803.63 & 13.81 & 0.2035 & 0.4447 & 0.9498 & 60.45 & 65.90 & 55.17 \\
    Animate-anything \cite{dai2023animateanything} & 352.79 & 21.02 & \textcolor{blue}{0.6490} & \textcolor{red}{0.1546} & 0.9668 & 84.31 & 87.13 & 80.75 \\
    \hline Ours & \textcolor{red}{226.13} & \textcolor{blue}{22.35} & \textcolor{red}{0.6431} & \textcolor{blue}{0.1637} & \textcolor{red}{0.9795} & \textcolor{red}{96.23} & \textcolor{red}{92.96} & \textcolor{red}{92.03} \\
    Ours(w/o $T$) & 293.47 & 20.79 & 0.5634 & 0.3416 & 0.9574 & 84.12 & 83.43 & 87.28 \\
    Ours(w/o motion fields) & 334.86 & 20.98 & 0.5866 & 0.3102 & 0.9321 & 80.67 & 81.65 & 82.84 \\
    Ours(w/o motion strength) & \textcolor{blue}{269.74} & 21.47 & 0.6115 & 0.1924 & 0.9682 & 92.74 & 90.23 & \textcolor{blue}{91.22} \\
    \hline
    \end{tabular}
\end{table*}

\begin{table*}
    \caption{Quantitative results of longer animation. The optimum value is distinguished by being highlighted in black.}
    \label{tab:qualitative_results_long}
    \begin{tabular}{l|cccccc|ccc}
    \hline & \multicolumn{6}{|c}{ Automatic Metrics} & \multicolumn{3}{c}{User Study} \\
    Method & FVD $\downarrow$ & PSNR $\uparrow$ & SSIM $\uparrow$ & LPISP $\downarrow$ & Tem-Cons $\uparrow$ & Time(s) $\downarrow$ & Mo-Coor $\uparrow$ & Vi-Co $\uparrow$ & Overall-Aes $\uparrow$\\
    \hline \hline Direct & 857.90 & 10.78 & 0.1534 & 0.4518 & 0.9066 & 86.65 & 56.21 & 61.32 & 50.89 \\
    Gen-L-Video \cite{wang2023gen} & 406.77 & 14.25 & 0.1764 & 0.2915 & 0.9346 & 162.43 & 88.31 & 82.76 & 79.91 \\
    FreeNoise \cite{qiu2023freenoise} & 337.86 & $\mathbf{19.93}$ & 0.5793 & 0.2487 & 0.9635 & 91.68 & 90.81 & 85.62 & 85.27 \\
    Ours & $\mathbf{298.62}$ & 19.49 & $\mathbf{0.5823}$ & $\mathbf{0.2135}$ & $\mathbf{0.9674}$ & $\mathbf{40.20}$ & $\mathbf{94.77}$ & $\mathbf{89.26}$ & $\mathbf{90.35}$ \\
    \hline
    \end{tabular}
\end{table*}

\noindent {\bf{Quantitative results.}}~ We quantitatively compared with open-sourced methods, including I2VGen \cite{2023i2vgenxl}, animating-pictures\cite{holynski2021animating}, SFS\cite{fan2023simulating}, animating-landscape \cite{2019Animating}, animate-anything\cite{dai2023animateanything}. We also work with the advanced commercial tools Gen-2 \cite{gen-2}, Pika Labs \cite{Pika} and Genmo \cite{Genmo}. Note that since commercial tools typically continue to iterate, we chose the March 15, 2024 version for comparison. To test the effect of our long video method, we compare it with the advanced video diffusion models' long video generation methods Gen-L-Video \cite{wang2023gen} and FreeNoise \cite{qiu2023freenoise}. Tab. \ref{tab:qualitative_results} displays the quantitative results. In comparison with the baselines, our model achieved the best scores, which substantiates the significant improvement in generation quality brought about by incorporating motion information control methods. The quantitative results for extended videos in Tab. \ref{tab:qualitative_results_long}, reveal that our method produces videos with better temporal consistency and visual coherence, demonstrating the efficacy of our approach for longer-duration video generation. Furthermore, our approach incorporates a phased inference process, which affords the capability to bypass unnecessary inferential steps during video extension. This peculiarity confers a substantial benefit in terms of reduced inference time, thereby enhancing the overall efficiency of our method.
\begin{figure}[h]
  \centering
  \includegraphics[width=\linewidth]{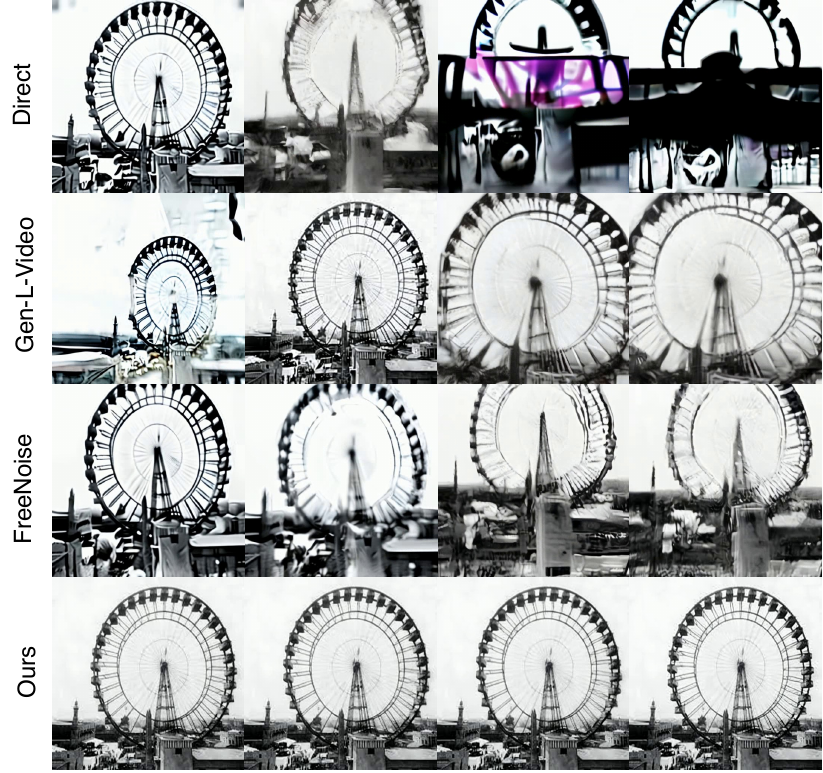}
  \caption{Qualitative results of longer video generation between baselines and our approach. The frames presented are 25, 50, 75, and 100 from the generated video. }
  \label{fig:longer_compare}
\end{figure}
\noindent {\bf{Qualitative results.}}~ We demonstrate visual examples in Fig. \ref{fig:compare_all}, comparing our method with open-source methods and commercial tools. While all evaluated methods are capable of producing seamless videos from static images, the outputs from Pika, SFS, Animating-landscape and Animating-pictures show the motion of wave that does not conform to physical laws in some scenarios. The outcome of Animate-anything reveals motion solely in the spray, while the bulk of the wave remains unexpectedly stationary. In the videos synthesized by Gen-2, there is unexpected movement not just within the waves but also on the beach, where tranquility should prevail. Meanwhile, Genmo's outputs are marred by jarring color transitions and visual distortions, evoking a sentiment of unreality. In contrast, our method yields videos that not only adhere more closely to the laws of physics but also maintain superior visual coherence. 

Fig. \ref{fig:longer_compare} displays the comparative analysis for the extended video generation method. The discrepancy between the training and inference  means that direct merging several short clips into a longer video can trigger substantial motion artifacts and cause spatial backgrounds to become blurry. This incongruity significantly undermines the overall quality of the synthesized video. For videos generated with a duration of 50 frames or fewer, FreeNoise, Gen-L-Video, and our method all exhibit commendable consistency in generation results. Nevertheless, as the length of the video continues to increase, the ability of FreeNoise and Gen-L-Video to maintain content constraints progressively weakens. Gen-L-Video employs the cross-frame attention method to interact between adjacent frames and the anchor frames, resulting in a lack of smooth temporal coherence at some junctions of video segments. FreeNoise incorporates a noise correlation scheduling strategy, yet it overlooks the interplay between structural and motion noise within the characteristics of image animation. Consequently, videos that are generated over extended lengths are prone to increasing distortion. Our approach deliberately distinguishes between content contours, background features, and motion intricacies during the process of noise reschedule. Through strategic decoupling of the inference phase and meticulous rescheduling of noise, we preserve the integrity of contours while synthesising motions. Consequently, this allows for a steady rotation of the Ferris wheel, devoid of any sudden shifts in the backdrop, and showcasing results with enhanced consistency.


\subsection{Ablation Study }
\begin{figure}[h]
  \centering
  \includegraphics[width=\linewidth]{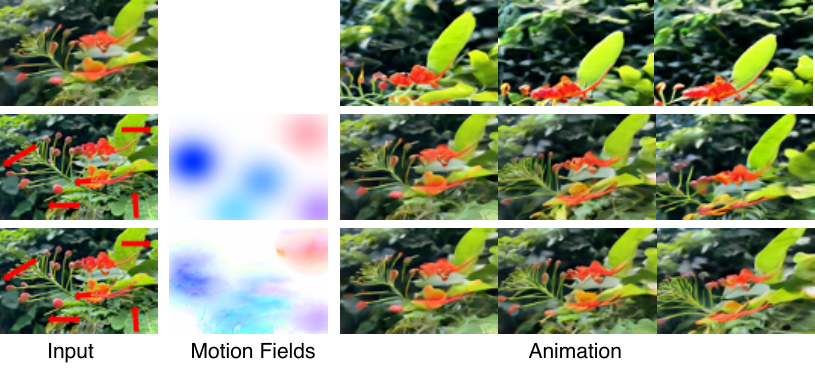}
  \caption{Ablation study about motion condition. The first row represents the outcome without motion field control(Ours w/o motion fields), the second row represents the outcome of controlling by dense motion field(Ours w/o $T$), and the third row represents the result of controlling by refine motion field.}
  \label{fig:ablation_flowers}
\end{figure}

\thispagestyle{empty}

\noindent {\bf{Motion condition.}}~ 
To investigate the influence of field constraints within our methodology, we examine three variants: 1) {\bf{Ours w/o $T$}}, the motion refinement model $T$ is omitted and dense optical flow is used as the control condition directly during the inference stage. 2) {\bf{Ours w/o motion fields}}, remove the entire optical flow motion field control module. 3) {\bf{w/o motion strength}}, remove the motion strength condition module. The outcomes of our quantitative assessments are detailed in Tab.\ref{tab:qualitative_results}, the performance of "Ours w/o motion fields" decreased significantly and Fig. \ref{fig:ablation_flowers} also corroborates that, in the absence of motion field constraints, flowers and green leaves exhibit uncontrolled movement, undermining the consistency of the content. When we remove the optical flow refinement model, the gap between the sparse track and the refine field adversely impact the quality of the generated video. Removing the motion strength module leads to unregulated global movement, which diminishes the visual appeal of the result. In contrast, our complete models effectively achieve animations of the highest quality and visual beauty.
\begin{figure}[h]
  \centering
  \includegraphics[width=\linewidth]{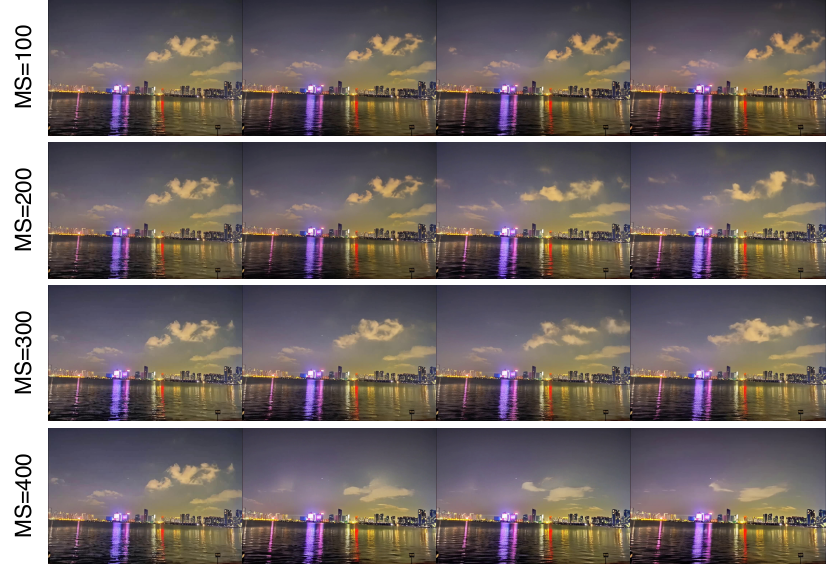}
  \caption{Ablation study on motion strength guidance. With the increment of motion strength, objects within the scene exhibit progressively higher speeds, yet they preserve synchronized temporal coordination throughout the process.}
  \label{fig:motion_strength}
\end{figure}

\noindent {\bf{Motion strength control.}}~ The effects of varying motion strength (MS) parameters are illustrated in Fig. \ref{fig:motion_strength}. We note that as MS values increase from $100$ to $200$, there are no substantial changes in the degree of motion depicted within the images. However, as MS escalates from $200$ to $400$, The dynamic elements of the animation have been significantly enhanced, with the fluctuation of the lake and the movement of clouds in the sky significantly faster. It is important to highlight that the object's motion adheres to a regular pattern of acceleration that is in line with the laws of physics, rather than changing erratically. The parts of the video that should not change (such as buildings) do not move, and  visual consistency is maintained during the acceleration process, eliminating any flickering in the imagery. Furthermore, visual consistency is maintained during the acceleration process, eliminating any flickering in the imagery. This observation serves as a testament to the efficacy of our motion strength control mechanism.
\begin{figure}[h]
  \centering
  \includegraphics[width=0.8\linewidth]{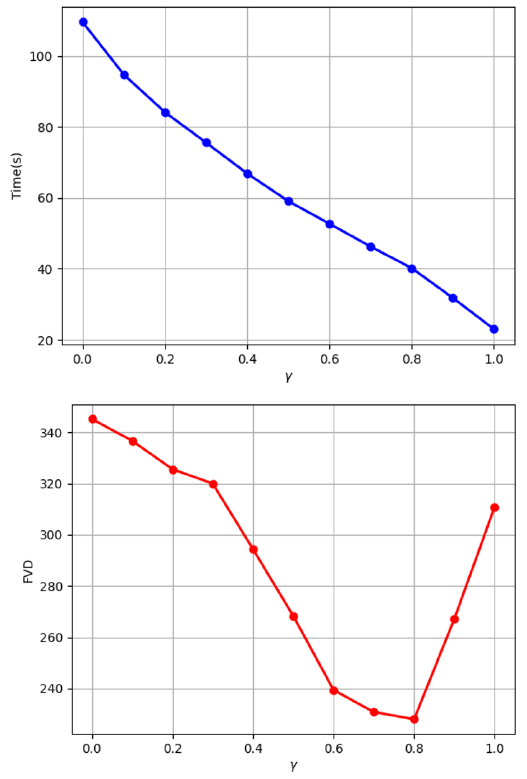}
  \caption{Ablation study about the FVD and longer animation inference time of hyperparameter $\gamma$. }
  \label{fig:ablation}
\end{figure}

\noindent {\bf{Phased inference hyperparameter $\gamma$.}}~ To explore the optimal timing for splitting the inference stage for the longer animation, as shown in Fig. \ref{fig:ablation}, we investigate the relationship between the parameter $\gamma$ and FVD and the long video generation inference times. As $\gamma$ incrementally increases, the quality of the generated video first improves and then decreases, reaching its optimal value within the range of approximately $0.7$ to $0.8$. Phased inference at the contour shaping stage leads to the alteration of contour information, which adversely affects the consistency of longer animation. By dividing the inference process during the motion detail enhancement phase, a significant amount of motion information is lost, subsequently impairing both the amplitude and variety of motion in extended video sequences. The inference time exhibits an inverse proportionality to the parameter $\gamma$. When $\gamma=1$, the long video inference completely degenerates into the replication of multiple short videos, and the inference time approaches the inference time of a single short video. Consequently, we establish  equilibrium among contour consistency, motion diversity, and inference efficiency by selecting $0.8$ as the value for $\gamma$.

\thispagestyle{empty}

\section{Conclusions}

In this work, we propose a method for generating longer dynamic videos from still images based on diffusion models, which controls the generated results with fine-grained motion fields and introduces an efficient method for extending long videos. Our method has demonstrated strong potential in terms of motion controllability and long video generation, overcoming the shortcomings of traditional methods that focus only on texture objects. However, using optical flow to describe the motion information of objects has limited capacity for content constraints. In future explorations, we will focus more on exploring methods that allow for more flexible multi-condition controls, such as sketch information, depth information, etc.

\thispagestyle{empty}



\bibliographystyle{ACM-Reference-Format}
\bibliography{arxiv}

\thispagestyle{empty}

\end{document}